\documentclass[acmlarge]{acmart}
\makeatletter
\newcommand{\confshort}{\acmConference@shortname}
\newcommand{\conffull}{\acmConference@name}
\newcommand{\confdate}{\acmConference@date}
\newcommand{\confloc}{\acmConference@venue}
\AtBeginDocument{
  \fancypagestyle{firstpagestyle}{
    \fancyhead{}%
    \fancyfoot[C]{}%
  }
  \fancyhf{}
  \fancyhead[LO]{\@headfootfont\shorttitle}%
  \fancyhead[RE]{\@headfootfont\@shortauthors}%
  \fancyhead[LE]{\@headfootfont\footnotesize \confshort, \confdate, \confloc}%
  \fancyhead[RO]{\@headfootfont\footnotesize \confshort, \confdate, \confloc}%
  \fancyfoot[C]{}%
}
\makeatother
\acmBooktitle{\conffull\@ (\confshort), \confdate, \confloc}

\AtBeginDocument{%
  }

\setcopyright{acmlicensed}
\copyrightyear{2026}
\acmYear{2026}
\acmConference[FAccT '26]{The 2026 ACM Conference on Fairness, Accountability, and Transparency}{June 25--28, 2026}{Montreal, Canada}
\acmBooktitle{The 2026 ACM Conference on Fairness, Accountability, and Transparency (FAccT '26), June 25--28, 2026, Montreal, QC, Canada}
\acmDOI{10.1145/3805689.3806448}
\acmISBN{979-8-4007-2596-8/2026/06}

\usepackage[utf8]{inputenc}
\usepackage{textgreek}

\begin{document}

%%
%% The "title" command has an optional parameter,
%% allowing the author to define a "short title" to be used in page headers.
\title[Recentering User Experience and Emotional Impact in the Evaluation of ASR Bias]{``This Wasn't Made for Me'': Recentering User Experience and Emotional Impact in the Evaluation of ASR Bias}

%%
%% The "author" command and its associated commands are used to define
%% the authors and their affiliations.
%% Of note is the shared affiliation of the first two authors, and the
%% "authornote" and "authornotemark" commands
%% used to denote shared contribution to the research.
\author{Siyu Liang}
% \authornote{Both authors contributed equally to this research.}
\email{liangsy@uw.edu}
% \orcid{0009-0002-2600-2509}
\author{Alicia Beckford Wassink}
% \authornotemark[1]
\email{wassink@uw.edu}
% \orcid{0000-0003-1805-8780}
\affiliation{%
  \institution{University of Washington}
  \city{Seattle}
  \state{Washington}
  \country{USA}
}

%%
%% By default, the full list of authors will be used in the page
%% headers. Often, this list is too long, and will overlap
%% other information printed in the page headers. This command allows
%% the author to define a more concise list
%% of authors' names for this purpose.
% \renewcommand{\shortauthors}{Trovato et al.}

%%
%% The abstract is a short summary of the work to be presented in the
%% article.
\begin{abstract}
While studies on bias in Automatic Speech Recognition (ASR) tend to focus on reporting error rates for speakers of underrepresented dialects, few investigate the human side of system bias: how do system failures shape users' lived experiences, how do users feel about and react to them, and what emotional toll do these repeated failures exact? We conducted user experience studies across four U.S. locations (Atlanta, Gulf Coast, Miami Beach, and Tucson) representing distinct English dialect communities. Our findings reveal that most participants feel that technologies fail to consider their cultural backgrounds and require constant adjustment to achieve basic functionality. Despite these experiences, participants maintain high expectations for ASR performance and express strong willingness to contribute to model improvement. Qualitative analysis of open-ended narratives exposes the deeper costs of these failures. Participants report frustration, annoyance, and feelings of inadequacy, yet the emotional impact extends beyond momentary reactions. Participants recognize that systems were not designed for them, yet often internalize failures as personal inadequacy despite this critical awareness. They perform extensive invisible labor, including code-switching, hyper-articulation, and emotional management, to make failing systems functional. Meanwhile, their linguistic and cultural knowledge remains unrecognized by technologies that encode particular varieties as standard while rendering others marginal. These findings demonstrate that algorithmic fairness assessments based on accuracy metrics alone miss critical dimensions of harm: the emotional labor of managing repeated technological rejection, the cognitive burden of constant self-monitoring, and the psychological toll of feeling inadequate in one's native language variety. Understanding ASR bias requires attending not just to error rates but centering evaluation on both the experiences and emotional impacts borne by affected users. 

\end{abstract}

%%
%% The code below is generated by the tool at http://dl.acm.org/ccs.cfm.
%% Please copy and paste the code instead of the example below.
%%
\begin{CCSXML}
<ccs2012>
   <concept>
       <concept_id>10010147.10010178.10010179.10010183</concept_id>
       <concept_desc>Computing methodologies~Speech recognition</concept_desc>
       <concept_significance>500</concept_significance>
       </concept>
   <concept>
       <concept_id>10003120.10011738.10011773</concept_id>
       <concept_desc>Human-centered computing~Empirical studies in accessibility</concept_desc>
       <concept_significance>500</concept_significance>
       </concept>
   <concept>
       <concept_id>10003120.10003121.10003122.10003334</concept_id>
       <concept_desc>Human-centered computing~User studies</concept_desc>
       <concept_significance>300</concept_significance>
       </concept>
   <concept>
       <concept_id>10003456.10010927.10003611</concept_id>
       <concept_desc>Social and professional topics~Race and ethnicity</concept_desc>
       <concept_significance>300</concept_significance>
       </concept>
 </ccs2012>
\end{CCSXML}

\ccsdesc[500]{Computing methodologies~Speech recognition}
\ccsdesc[500]{Human-centered computing~Empirical studies in accessibility}
\ccsdesc[300]{Human-centered computing~User studies}
\ccsdesc[300]{Social and professional topics~Race and ethnicity}

%%
%% Keywords. The author(s) should pick words that accurately describe
%% the work being presented. Separate the keywords with commas.
\keywords{Automatic Speech Recognition, ASR, fairness, algorithmic bias, dialect bias, user experience, emotional impact, voice technology, African American English, Chicano English, Native American English, Gulf Coast English}

% \received{13 January 2026}
% \received[revised]{12 March 2009}
% \received[accepted]{5 June 2009}

%%
%% This command processes the author and affiliation and title
%% information and builds the first part of the formatted document.
\maketitle

\section{Introduction}

Automatic Speech Recognition (ASR) systems have become deeply embedded in everyday technologies, from smartphones and smart speakers to customer service systems and accessibility tools. These systems promise hands-free convenience and natural interaction, yet mounting evidence demonstrates substantial performance disparities across dialects and accents. For speakers of non-mainstream language varieties, technical failures occur systematically and frequently, carrying consequences that extend far beyond mere inconvenience. 

Prior research has documented these performance gaps extensively. Commercial ASR systems exhibit higher word error rates for certain groups of English speakers such as African American English speakers, Spanish-influenced English speakers, and Indigenous language speakers compared to speakers of dominant English varieties \cite{koenecke_racial_2020, wassink_uneven_2022, martin_understanding_2020}. These disparities stem primarily from training data imbalances and linguistic features characteristic of non-dominant varieties that systems fail to accommodate \cite{tatman_gender_2017, feng_towards_2024, ngueajio_hey_2022}. These studies have established that bias exists and quantified its magnitude through error metrics, phonetic analyses, and controlled evaluations. These performance disparities reflect broader patterns of algorithmic discrimination documented across AI systems \cite{noble_algorithms_2018, benjamin_race_2019}.

Yet while the documentation of ASR bias continues to accumulate, significantly less attention has been paid to the human side of system bias. How do system failures shape users' lived experiences? How do users feel about and react to repeated misrecognition? What emotional toll do these failures exact over time? How do experiences vary across different dialect communities facing similar technical failures? And critically, how do these experiences shape users' relationships with voice technologies and their willingness to contribute to improving biased systems? 

This study addresses these questions through a mixed-methods investigation of user experiences across four U.S. dialect communities: African American speakers in Atlanta, Georgia; Gulf Coast Creole and French speakers in Mississippi and Louisiana; Hispanic and Latine Caribbean speakers in Miami Beach, Florida, and Native American speakers in Tucson, Arizona. Through quantitative analysis of structured questionnaires and qualitative analysis of open-ended narrative responses, we move beyond performance metrics to examine the lived experiences and emotional impacts of algorithmic bias: the frustrations users articulate, the accommodations they make, their understandings of why systems fail them, and the complex decisions they face about whether to persist with, abandon, or contribute towards improving technologies that exclude them. 

Our findings reveal that participants maintain high expectations for ASR performance and express overwhelming willingness to contribute to model improvement, yet most report that technologies fail to consider their cultural backgrounds and require constant adjustment to achieve basic functionality. Participants report frustration, annoyance, and feelings of inadequacy, yet the emotional impact extends beyond these momentary reactions. Qualitative analysis exposes deeper psychological costs: participants recognize that systems were not designed for them (``this wasn't made for me''), yet internalize failures as personal inadequacy despite this critical awareness. They perform extensive invisible labor, including code-switching, hyper-articulation, and emotional management, to make failing systems functional. Their linguistic and cultural knowledge remains unrecognized by technologies that encode particular varieties as standard while rendering others marginal. This tension between understanding exclusion and continuing to accommodate failing systems illuminates the complex situation of users navigating technologies that have become increasingly unavoidable yet remain systematically unreliable for their speech. 

This research contributes to fairness and accountability scholarship by recentering user experience and emotional impact in evaluations of algorithmic bias. We demonstrate that fairness assessments based on accuracy metrics alone miss critical dimensions of harm: the emotional labor of managing repeated technological rejection, the cognitive burden of constant self-monitoring, and the psychological toll of feeling inadequate in one's native language. These invisible costs are central to experiences of algorithmic exclusion yet remain uncaptured in technical evaluations. We document how users interpret and respond to systematic technological failure, revealing sophisticated critical awareness alongside internalized self-blame, and exposing forms of harm that performance metrics fail to capture. Understanding ASR bias requires attending not just to error rates but to how bias is experienced and felt by affected users. 

\section{Related Work}

Our work sits at the intersection of ASR bias analysis, user experience research on algorithmic harm, sociolinguistics, and critical frameworks for understanding linguistic justice in AI systems. We review three strands of research central to our research questions: (1) documented performance disparities and their causes, (2) user experiences of algorithmic bias and technological invisibility, and (3) participatory responses to biased systems and design justice. 

\subsection{ASR Performance Disparities Across Dialects}

Studies have consistently documented performance disparities in commercial ASR systems across racial, ethnic, and dialectal lines. Koenecke et al. \cite{koenecke_racial_2020} found that major commercial systems from Amazon, Apple, Google, IBM, and Microsoft exhibited higher word error rates for African American speakers compared to white speakers, with disparities particularly pronounced for speakers of African American English (AAE). These disparities stem primarily from training data imbalances, with ASR models predominantly trained on Standard American English from speakers who are disproportionately white \cite{dorn_dialect-specific_2019}. 

Recent work has expanded evaluation to multiple dialect varieties. Wassink et al. \cite{wassink_uneven_2022} analyzed ASR errors across four ethnicity-related sociolects from the Pacific Northwest, demonstrating differential error rates: Caucasian American English exhibited lowest error rates, followed by African American English, Yakama English, and Chicano English. Harris et al. \cite{harris_modeling_2024} evaluated ASR performance across four U.S. dialects (Standard American English, AAE, Chicano English, and Spanglish), finding intersectional gender-dialect biases where systems performed worst for female speakers of minority dialects. Martin and Tang \cite{martin_understanding_2020} showed how ASR systems fail to recognize the habitual ``be'' construction characteristic of AAE, a morphosyntactic feature with systematic grammatical function that systems misinterpret or ignore. Choe et al. \cite{choe_language-specific_2022} examined how speakers' native language phonologies predict specific categories of ASR errors across World Englishes. Approaches have also focused on different ways to develop systematic methods for measuring and analyzing ASR bias. For example, Feng et al. \cite{feng_quantifying_2021} analyzed specific phonemes to identify sources of bias, while Liu et al. \cite{liu_towards_2021} developed fairness measurement frameworks using demographically diverse datasets. However, critical analysis of this research literature itself reveals persistent conceptual limitations, including misconceptions about accent as something only certain speakers possess \cite{prinos_speaking_2024}. 

This body of work establishes that performance disparities are substantial, systematic, and attributable to specific phonological or morphosyntactic features characterizing non-dominant varieties. However, these studies focus primarily on \textit{documenting} failure rates rather than examining \textit{user experiences} of failure or how communities respond when faced with persistent technological exclusion---the gaps our work addresses.

\subsection{User Experiences of Algorithmic Bias}

Emerging research examines how users experience algorithmic bias emotionally and psychologically. Mengesha et al. \cite{mengesha_i_2021} investigated African American users' experiences with ASR errors through interviews and surveys, finding that participants perceived errors as culturally insensitive and felt the technology was not designed for them. Their qualitative work documented frustration, resignation, and perceptions of technological exclusion, providing crucial evidence that ASR bias creates psychological impacts beyond mere inconvenience. However, questions remain about how these feelings of exclusion interact with users' ongoing decisions to use, accommodate, or abandon failing technologies. Recent work has extended this analysis to AI voice services more broadly. Michel et al. \cite{michel_its_2025} examined accent bias and digital exclusion in synthetic AI voice services (Speechify and ElevenLabs), finding parallel patterns of users feeling unrepresented by technologies that fail to accommodate their speech patterns.

Research on intersectional invisibility in organizational contexts provides a theoretical framework for understanding these experiences. Bhattacharyya and Berdahl \cite{bhattacharyya_you_2023} documented how women of color in workplaces experience systematic oversight and misrecognition at the intersection of multiple marginalized identities, identifying four forms of invisibility: erasure (being overlooked), homogenization (being confused with others), whitening (having cultural identity ignored), and exoticization (being treated as exotic other). Critically, these invisibility experiences create distinct affective reactions: ambiguous events (erasure, whitening) elicit shame and self-blame, while less ambiguous events (homogenization, exoticization) elicit anger directed at perpetrators. This dual pathway suggests that even when structural causes are visible, internalized responses may persist. This work builds on broader scholarship documenting how marginalized groups experience systematic misrecognition in institutional contexts, with implications for how users interpret technological failures \cite{rickford_language_2016}.

We theorize that similar dynamics operate in human-technology interaction. When ASR systems consistently fail to recognize certain speakers, these failures may create \textit{technological invisibility}---experiences of being unseen, misrecognized, or overlooked by systems. An open question is whether users maintain critical awareness that systems were not designed for them while simultaneously internalizing failure as personal inadequacy, and how these contradictory interpretations coexist in everyday technological practice. 

\subsection{Technological Labor and Participatory Tensions}
\label{sec:tech_labor}

When speakers of non-mainstream varieties must accommodate dominant norms to participate in technological systems, this accommodation constitutes a form of invisible labor: cognitive effort to modify natural speech, emotional management when systems fail, and temporal costs of repetition and troubleshooting \cite{mengesha_i_2021}.  We draw upon the long history of accommodation research conducted in sociolinguistics and social psychology. Accommodation in the speech domain refers to the conscious or unconscious alterations a talker makes to their linguistic or paralinguistic self-presentation, either in response to interpersonal factors (such as the desire to enhance perceived social proximity or social attractiveness), or in response to a change in topic or interlocutor, or of a desire to control the persona that they project (e.g., to convey humor or playfulness)\cite{Giles_Powesland_1975, Bell_1984, Eckert_2000, Coupland_2001}.  Collectively, these are factors influencing style-shifting. Crucially for our purposes, there is neuro-linguistic research indicating that, particularly in diglossic communities where social norms dictate that  each dialect be relegated to a different social domain, bidialectal speakers may incur cognitive switching costs that monolinguals do not, reflecting the need to inhibit the activation of the variety that is inappropriate for a given setting \cite{Kirk_et_al_2021, Vorwerg_et_al_2019}.   We use the term \textit{invisible labor} to encompass three interrelated dimensions. First, drawing on Hochschild's \cite{hochschild_managed_2012} concept of \textit{emotional labor}---the work of managing one's feelings to fulfill external expectations---we recognize that ASR users must suppress frustration, self-blame, and alienation to continue engaging with failing systems. Second, extending this concept through Baker-Bell's \cite{baker-bell_linguistic_2020} framework of internalized linguistic racism, we identify a specifically \textit{linguistic} dimension: the cognitive and emotional work of suppressing one's natural speech variety to conform to a system's narrow expectations, what we term \textit{linguistic labor}. Third, following Cunningham et al. \cite{cunningham_understanding_2024}, we attend to the practical accommodation strategies---repetition, code-switching, hyper-articulation---that consume time and attention. This labor remains invisible to developers and privileged-dialect users for whom ASR ``just works,'' yet represents a significant and inequitable burden for those whose speech falls outside training distributions. Pospisil et al. \cite{pospisil_totally_2022} find that about 76 percent of those who find the labor and concerns to be excessive will never return to the use of voice assistants.  Our study operationalizes this concept through systematic documentation of coping strategies, examining what accommodations users report making and at what frequency across different dialect communities. The accommodation burden placed on speakers of non-dominant varieties reflects broader patterns of linguistic discrimination, wherein certain ways of speaking are systematically devalued while others are treated as neutral or standard \cite{lippi-green_english_2012}. 

Design justice frameworks argue that equitable technologies require centering marginalized communities throughout design processes \cite{costanza-chock_design_2020}. Costanza-Chock articulates principles emphasizing that those most affected by design decisions should lead design processes, that communities hold expertise about their own experiences, and that design should dismantle rather than reproduce structural inequalities. Applied to ASR, these principles suggest that developing dialect-inclusive systems requires sustained partnership with speakers of underrepresented varieties, not merely collecting their voice data.

Yet tensions emerge when communities experiencing bias are asked to contribute to fixing biased systems. Harrington et al. \cite{harrington_deconstructing_2019} examine community-based collaborative design with Black and Brown communities, identifying tensions between researcher and community goals and emphasizing the importance of compensation, shared decision-making authority, and recognition of community expertise. When voice sample collection occurs without these protections, questions arise about equitable participation in AI development. Recent work has begun addressing these tensions in ASR contexts specifically. Hutchinson et al. \cite{hutchinson_designing_2025} document participatory approaches to developing speech technologies for Australian Aboriginal English, highlighting both opportunities and risks when developing ASR for marginalized language communities. A critical question emerges: if users recognize that ASR was not designed for them and experience repeated failures, what motivates continued willingness to participate in system improvement? This question sits at the intersection of critical awareness and ongoing technological engagement. Our study examines whether communities experiencing significant harm nevertheless remain willing to contribute to model improvement, and how this willingness relates to their experiences of exclusion and their understanding of why systems fail them. 

\subsection{Positioning This Work}

This work reframes the study of ASR bias by shifting attention from system-centric performance metrics to the social and experiential dimensions of human–technology interaction. Rather than asking only how recognition accuracy varies across dialects, we ask how persistent misrecognition is understood, navigated, and emotionally processed by users whose speech falls outside dominant training distributions. By situating ASR failures within users’ everyday practices, expectations, and interpretive frameworks, this study treats bias not solely as a technical property of models, but as a lived condition shaped by design choices, institutional norms, and unequal distributions of accommodation labor. In doing so, we bridge technical research on ASR disparities with human-centered and justice-oriented approaches to algorithmic accountability, emphasizing the importance of user experience as a central site for evaluating fairness in voice technologies.

\section{Methodology}

\subsection{Research Design}

Our study investigates user experiences with English Automatic Speech Recognition (ASR) technologies across four geographically and linguistically diverse communities in the United States. Our methods combine structured questionnaires with open-ended responses to capture both quantitative patterns and qualitative experiences of ASR system failures and their psycho-emotional impacts. 

Data collection occurred between July, 2023 and August, 2025 for four sites: Atlanta, Georgia; the Gulf Coast region (Mississippi and Louisiana); Miami Beach, Florida; and Tucson, Arizona. These communities represent systematically underrepresented varieties with well-documented linguistic structures distinct from Standard American English \cite{green_african_2002, blodgett_demographic_2016, fought_chicano_2003, leap_american_1993, klingler_if_2003}.

\subsection{Participants}
Recruitment was accomplished by word-of-mouth, digital ad postings (Twitter, FaceBook/Meta, Craigslist), and flyers posted in public places (university campuses, laundromats, grocery stores, bookstores, churches, community centers). We recruited a total of 494 participants across the four sites; of these, 215 of them met our inclusion criteria (target linguistic and demographic background) and provided near-complete survey responses. These 215 participants form the basis of this study.  All participants self-identified as speakers of underrepresented dialects in the United States and reported regular use of voice-enabled technologies. Eight members of the research team (2-6 per location) conducted participant recruitment.  Recruitment strategies varied by site to reach distinct dialect communities: Atlanta participants (n=86) were recruited at local community centers and a conference of black engineers; Gulf Coast participants (n=61) were recruited from the community by word-of-mouth; Miami Beach participants (n=43) were recruited from the student and staff body at Florida International University; and Tucson participants (n=25) were recruited through the Pascua Yaqui Cultural Center and the intertribal Native American Student Affairs center at the University of Arizona. 

Table~\ref{tab:demographics} presents an overview of the demographic composition of the participants. The Atlanta sample (n=86) consisted primarily of Black/African American participants from the Southeast. The Gulf Coast sample (n=61) included Black/African American participants and French Creole speakers from Mississippi and Louisiana. The Miami Beach sample (n=43) included Caribbean Hispanic/Latine participants, Black Caribbean, and Black/African American participants from Florida. The Tucson sample (n=25) consisted primarily of Native American participants from Arizona, including speakers of Yaqui, Navajo, Tohono O'odham, and other Indigenous backgrounds.

% The total of 494 participants is based on Atlanta 99, Gulf Coast 327, Miami Beach 43, and Tucson 25
% The following Atlanta participants contain empty reponses
% CD24-009, CD24-012, CD24-033, CD24-039, CD24-065, CD24-066, CD24-067, CD24-068, CD24-085, CD24-088, CD24-089, CD24-182, CD24-223

\begin{table}[h]
\caption{Participant Demographics by Research Site}
\label{tab:demographics}
\begin{tabular}{lcccc}
\toprule
\textbf{Characteristic} & \textbf{Atlanta} & \textbf{Gulf Coast} & \textbf{Miami Beach} & \textbf{Tucson} \\
 & (n=86) & (n=61) & (n=43) & (n=25) \\
\midrule
\multicolumn{5}{l}{\textit{Gender Identity}} \\
\quad Women & 48 (56\%) & 20 (33\%) & 31 (72\%) & 19 (76\%) \\
\quad Men & 37 (43\%) & 38 (62\%) & 11 (26\%) & 6 (24\%) \\
\quad Non-binary & 1 (1\%) & 1 (2\%) & 0 (0\%) & 0 (0\%) \\
\quad Trans woman & 0 (0\%) & 1 (2\%) & 0 (0\%) & 0 (0\%) \\
\quad Prefer not to state & 0 (0\%) & 1 (2\%) & 1 (2\%) & 0 (0\%) \\
\midrule
\multicolumn{5}{l}{\textit{Age (years)}} \\
\quad Range & 19--60 & 18--68 & 18--63 & 18--54 \\
\quad Mean (SD) & 29.9 (8.5) & 33.6 (12.1) & 23.4 (11.2) & 25.8 (10.2) \\
\bottomrule
\end{tabular}
\vspace{2mm}
\end{table}

\subsection{Data Collection}

\subsubsection{Questionnaire Design}

We developed a questionnaire investigating participants' experiences with voice-enabled technologies, their perceptions of ASR system performance, and the psycho-emotional impacts of recognition failures. The questionnaire included both structured multiple-choice items and open-ended narrative prompts. 

Key questionnaire topics included:
\begin{itemize}
\item Demographic information (age, location, ethnicity, gender identity)
\item Self-description of speech characteristics (Q1) and speech challenges (Q4)
\item Types of voice-enabled technologies used (Q5--Q7) and purposes of use (Q6a, Q8a)
\item Frequency of use for different purposes (Q12a--f: 5-point Likert scale)
\item Discontinued use and reasons (Q10: open-ended)
\item Memorable experiences with ASR technologies (Q11: open-ended)
\item Specific instances of dialect or accent-related recognition failures (Q13: open-ended)
\item Types of challenges experienced (Q14: multiple-choice with options including cultural/ethnic/racial exclusion, contextual misunderstanding, physical effort requirements, self-consciousness)
\item Coping strategies employed when systems fail (Q15a: multiple-choice with options including repetition, speech modification, enunciation, speaking louder/slower, imitation, manual task completion, asking others for help, technology abandonment)
\item Emotional responses to system failures (Q16: open-ended)
\item Expectations for ASR accuracy (Q17: 5-point scale from ``Always [100\%]'' to ``Never [0--20\%]'')
\item Willingness to provide voice samples for system improvement (Q18a: Yes/Maybe/No)
\item Willingness to pay for improved services (Q19: multiple payment tier options including compensation preference)
\end{itemize}

Questionnaire items were developed using a two-part process. Prompts originally prepared by the second author were reviewed and updated in collaboration with a UX researcher (a member of the team) with experience using culturally-informed community-based participatory research design frameworks for fairness in technology (CBPR) \cite{harrington_deconstructing_2019}. The questionnaire was then brought to the Seattle Black Community Advisory Board (B-CAB), who provided further feedback on item wording, overall content, and impact of the questionnaire as well as the alignment of the project with the B-CAB's own goals to advance technological fairness in the local community.  B-CAB feedback was incorporated into the questionnaire. 

The full questionnaire can be found in the appendix. The Atlanta collection was the first to occur. Afterward, the interview and survey were modified slightly to increase the likelihood that question wording would elicit the information of interest. We will mention  below where the Atlanta questionnaire differed slightly from the other locations' surveys.

\subsubsection{Data Collection Procedures}

Study procedures were approved by the University of Washington Human Subjects Division (Review\#: 00019648). Two procedures were used for the collection of data.  First, a semi-structured interview was delivered in-person and audio-recorded. Following consenting procedures, respondents were invited to respond verbally to an interviewer who delivered all questions orally.  Second, we offered the choice to all respondents to participate online via a self-paced survey created using the JotForm platform (jotform.com). The survey appeared only after online consent was given.  The survey included radio-button, multiple-choice, short-form text entry and long-form text entry as well as audio widget response options, as appropriate to the question.  All participants, regardless of collection method, responded to the same questionnaire, enabling cross-site comparison for both quantitative and qualitative analyses. 

\subsection{Analytical Approach}

\subsubsection{Quantitative Analysis}

We report descriptive statistics (frequencies, percentages) for all structured questionnaire items to characterize response patterns within and across sites. For multiple-selection items (Q14: challenges experienced; Q15a: coping strategies employed), participants could select all applicable options, resulting in non-mutually exclusive categories. Percentages therefore reflect the proportion of participants who endorsed each option and may sum to greater than 100\%. Comparisons across sites are reported descriptively given the non-probability sampling approach and the distinct demographic characteristics of each dialect community. Where notable differences emerged between sites, we interpret these in light of the specific linguistic and cultural contexts of each community. 

\subsubsection{Qualitative Analysis}

We conducted qualitative analysis of open-ended responses to three key questions: memorable experiences with ASR technologies (Q11), specific instances of dialect-related recognition challenges (Q13), and emotional responses to system failures (Q16). Several major themes emerged from this analysis, representing distinct dimensions of how participants experience and respond to ASR bias. These themes capture both surface-level impacts (frustration with errors, need for repetition) and deeper psychological dimensions (internalized blame despite critical awareness of exclusion, feelings of linguistic and cultural invisibility).

The qualitative analysis approach taken was an adaptation of thematic content analysis as established within Grounded Theory \cite{glaser_strauss_1967}. Audio-recorded responses to survey questions were automatically transcribed, then human-corrected. Corrected transcripts were next subjected to iterative coding passes to enable highlighting of common themes arising across the dialect databases and themes related to the research questions of the study. In the first pass, two team members per dialect coded utterances using a list of 14 generalized tags defined in a conversational coding guide designed to address the research questions of the study. Inter-coder reliability was assessed using Krippendorff's alpha ($\alpha \geq 0.80$). Several steps were made to improve inter-coder agreement. Weekly team meetings were used to resolve ambiguous cases, discuss code definitions, highlight interesting responses and only if necessary, update the codebook with parsimony. Coders then re-coded their materials, if the threshhold for agreement was not yet achieved. In this paper, we focus on responses coded as `feelings' (FEL), `presuppositions' (PSP), `narratives' (NARR), and `personifications' (PERS).

Participant quotes are identified by site and unique participant ID (e.g., [Atlanta] CD24-055) to enable verification while protecting anonymity. When quotes contained implicit identifying information, minor details were altered to preserve confidentiality without changing the meaning substantively.  

\section{Results}

We present findings below from our investigation of user experiences with ASR technologies. Since some participants could choose to skip some questions, response rates varied by question. Therefore, sample sizes differ slightly across questions and are reported accordingly in the respective tables. 

\subsection{Quantitative Findings}

\subsubsection{Types of Challenges Experienced}
Participants identified multiple types of challenges when using ASR technologies (Table~\ref{tab:q14_challenges}). Cultural and ethnic exclusion was reported by a majority of participants overall (53.5\%), with notably higher rates among Tucson participants (72.0\%) and Miami Beach participants (62.8\%) compared to Gulf Coast (52.5\%) and Atlanta (44.2\%) participants.

Contextual misunderstanding---voice assistants taking speech out of context or performing unintended actions---was reported by 51.2\% of participants overall, with 76.0\% of Tucson participants and 69.8\% of Miami Beach participants reporting the issue, compared to 47.5\% of Gulf Coast and 37.2\% of Atlanta participants.

Physical effort requirements showed a similar pattern, with Tucson (60.0\%) and Miami Beach (39.5\%) participants reporting higher rates than Gulf Coast (37.7\%) and Atlanta (27.9\%) participants. Self-consciousness about speech when interacting with voice technologies was reported by 44.0\% of Tucson participants, compared to 36.1\% of Gulf Coast, 25.6\% of Miami Beach, and 17.4\% of Atlanta participants.

\begin{table}[h]
\centering
\caption{Types of Challenges Experienced by Site (Q14; Multiple Selections Allowed)}
\label{tab:q14_challenges}
\begin{tabular}{lcccccccccc}
\toprule
 & \multicolumn{2}{c}{Atlanta} & \multicolumn{2}{c}{Gulf Coast} & \multicolumn{2}{c}{Miami Beach} & \multicolumn{2}{c}{Tucson} & \multicolumn{2}{c}{Total} \\
 & \multicolumn{2}{c}{(n=86)} & \multicolumn{2}{c}{(n=61)} & \multicolumn{2}{c}{(n=43)} & \multicolumn{2}{c}{(n=25)} & \multicolumn{2}{c}{(N=215)} \\
\cmidrule(lr){2-3} \cmidrule(lr){4-5} \cmidrule(lr){6-7} \cmidrule(lr){8-9} \cmidrule(lr){10-11}
Challenge Type & n & \% & n & \% & n & \% & n & \% & n & \% \\
\midrule
Cultural/ethnic/racial exclusion & 38 & 44.2 & 32 & 52.5 & 27 & 62.8 & 18 & 72.0 & 115 & 53.5 \\
Contextual misunderstanding & 32 & 37.2 & 29 & 47.5 & 30 & 69.8 & 19 & 76.0 & 110 & 51.2 \\
Physical effort required & 24 & 27.9 & 23 & 37.7 & 17 & 39.5 & 15 & 60.0 & 79 & 36.7 \\
Self-conscious about speech & 15 & 17.4 & 22 & 36.1 & 11 & 25.6 & 11 & 44.0 & 59 & 27.4 \\
\bottomrule
\end{tabular}
\end{table}

\subsubsection{Coping Strategies}
When ASR systems failed to recognize their speech, participants reported employing a variety of coping strategies (Table~\ref{tab:q15_strategies}). Repetition was the most common strategy overall (82.8\%), with consistently high rates across all sites ranging from 75.4\% in the Gulf Coast to 87.2\% in Atlanta.

Speech modification was a prevalent coping mechanism. In Atlanta, where participants were offered a more open-ended response option, 58.1\% reported changing the way they speak or code-switching. At other sites, where more granular options were available, participants reported slowing down speech (Gulf Coast 68.9\%, Miami Beach 72.1\%, Tucson 76.0\%), speaking louder (Gulf Coast 59.0\%, Miami Beach 83.7\%, Tucson 88.0\%), enunciating more (Gulf Coast 47.5\%, Miami Beach 83.7\%, Tucson 84.0\%), and changing words or phrases (Gulf Coast 54.1\%, Miami Beach 60.5\%, Tucson 72.0\%).

Technology abandonment strategies were also prevalent. Complete discontinuation of ASR use was reported by 49.8\% of participants overall, with Tucson showing the highest rate (80.0\%), followed by Miami Beach (55.8\%), Atlanta (47.7\%), and Gulf Coast (36.1\%). Manual task completion was common in the Gulf Coast (59.0\%), Miami Beach (76.7\%), and Tucson (80.0\%). Social workarounds included imitating the voice assistant's speech patterns (26.5\% overall) and asking another person to operate the device (21.9\% overall).

\begin{table}[h]
\centering
\caption{Coping Strategies Employed by Site (Q15a; Multiple Selections Allowed)}
\label{tab:q15_strategies}
\begin{tabular}{lcccccccccc}
\toprule
 & \multicolumn{2}{c}{Atlanta} & \multicolumn{2}{c}{Gulf Coast} & \multicolumn{2}{c}{Miami Beach} & \multicolumn{2}{c}{Tucson} & \multicolumn{2}{c}{Total} \\
 & \multicolumn{2}{c}{(n=86)} & \multicolumn{2}{c}{(n=61)} & \multicolumn{2}{c}{(n=43)} & \multicolumn{2}{c}{(n=25)} & \multicolumn{2}{c}{(N=215)} \\
\cmidrule(lr){2-3} \cmidrule(lr){4-5} \cmidrule(lr){6-7} \cmidrule(lr){8-9} \cmidrule(lr){10-11}
Strategy & n & \% & n & \% & n & \% & n & \% & n & \% \\
\midrule
Repeat utterance & 75 & 87.2 & 46 & 75.4 & 36 & 83.7 & 21 & 84.0 & 178 & 82.8 \\
Code-switch/modify speech\textsuperscript{a} & 50 & 58.1 & -- & -- & -- & -- & -- & -- & -- & -- \\
Change words/phrases\textsuperscript{b} & -- & -- & 33 & 54.1 & 26 & 60.5 & 18 & 72.0 & 77 & 59.7 \\
Slow down speech\textsuperscript{b} & -- & -- & 42 & 68.9 & 31 & 72.1 & 19 & 76.0 & 92 & 71.3 \\
Speak louder\textsuperscript{b} & -- & -- & 36 & 59.0 & 36 & 83.7 & 22 & 88.0 & 94 & 72.9 \\
Enunciate more\textsuperscript{b} & -- & -- & 29 & 47.5 & 36 & 83.7 & 21 & 84.0 & 86 & 66.7 \\
Stop using technology & 41 & 47.7 & 22 & 36.1 & 24 & 55.8 & 20 & 80.0 & 107 & 49.8 \\
Complete task manually\textsuperscript{b} & -- & -- & 36 & 59.0 & 33 & 76.7 & 20 & 80.0 & 89 & 69.0 \\
Imitate voice assistant & 14 & 16.3 & 16 & 26.2 & 20 & 46.5 & 7 & 28.0 & 57 & 26.5 \\
Ask another person & 9 & 10.5 & 13 & 21.3 & 16 & 37.2 & 9 & 36.0 & 47 & 21.9 \\
\bottomrule
\multicolumn{11}{l}{\textsuperscript{a}\footnotesize{Atlanta used an earlier survey version with this broader category instead of granular speech modification options.}} \\
\multicolumn{11}{l}{\textsuperscript{b}\footnotesize{Option not available for Atlanta; percentage calculated from Gulf Coast, Miami Beach, and Tucson only (n=129).}}
\end{tabular}
\end{table}

\subsubsection{Expectations for ASR Performance}

Most participants expected ASR systems to understand their speech correctly at least 75\% of the time (Table~\ref{tab:q17_expectations}). Across all sites, 53.7\% of participants expected ASR to work ``often'' (75\% of the time), and 27.6\% expected it to work ``always'' (100\% of the time). Only 15.0\% expected ASR to work only ``sometimes'' (50\%), and 3.7\% expected it to work ``rarely'' (25\%) or never. These high expectations---with over 80\% anticipating at least 75\% accuracy---contrast sharply with the challenges and coping strategies reported above, suggesting a substantial gap between user expectations and actual ASR performance for these dialect communities. 

\begin{table}[h]
\centering
\caption{Expectations for ASR Accuracy by Site (Q17)}
\label{tab:q17_expectations}
\begin{tabular}{lcccccccccc}
\toprule
 & \multicolumn{2}{c}{Atlanta} & \multicolumn{2}{c}{Gulf Coast} & \multicolumn{2}{c}{Miami Beach} & \multicolumn{2}{c}{Tucson} & \multicolumn{2}{c}{Total} \\
 & \multicolumn{2}{c}{(n=85)} & \multicolumn{2}{c}{(n=61)} & \multicolumn{2}{c}{(n=43)} & \multicolumn{2}{c}{(n=25)} & \multicolumn{2}{c}{(N=214)} \\
\cmidrule(lr){2-3} \cmidrule(lr){4-5} \cmidrule(lr){6-7} \cmidrule(lr){8-9} \cmidrule(lr){10-11}
Expected Accuracy & n & \% & n & \% & n & \% & n & \% & n & \% \\
\midrule
Always (100\%) & 28 & 32.9 & 15 & 24.6 & 9 & 20.9 & 7 & 28.0 & 59 & 27.6 \\
Often (75\%) & 43 & 50.6 & 32 & 52.5 & 28 & 65.1 & 12 & 48.0 & 115 & 53.7 \\
Sometimes (50\%) & 12 & 14.1 & 11 & 18.0 & 4 & 9.3 & 5 & 20.0 & 32 & 15.0 \\
Rarely (25\%) & 2 & 2.4 & 3 & 4.9 & 1 & 2.3 & 1 & 4.0 & 7 & 3.3 \\
Never (0--20\%) & 0 & 0.0 & 0 & 0.0 & 1 & 2.3 & 0 & 0.0 & 1 & 0.5 \\
\bottomrule
\end{tabular}
\end{table}

\subsubsection{Willingness to Contribute Voice Samples}

Despite experiencing significant challenges with ASR systems, a substantial majority of participants (75.5\%) expressed willingness to contribute voice samples to help improve recognition accuracy for diverse dialects (Table~\ref{tab:q18_samples}). An additional 17.9\% indicated they might contribute, while only 6.6\% declined. Willingness was highest in Miami Beach (88.1\%) and Gulf Coast (85.2\%), followed by Tucson (76.0\%) and Atlanta (61.9\%). 

\begin{table}[h]
\centering
\caption{Willingness to Provide Voice Samples by Site (Q18a)}
\label{tab:q18_samples}
\begin{tabular}{lcccccccccc}
\toprule
 & \multicolumn{2}{c}{Atlanta} & \multicolumn{2}{c}{Gulf Coast} & \multicolumn{2}{c}{Miami Beach} & \multicolumn{2}{c}{Tucson} & \multicolumn{2}{c}{Total} \\
 & \multicolumn{2}{c}{(n=84)} & \multicolumn{2}{c}{(n=61)} & \multicolumn{2}{c}{(n=42)} & \multicolumn{2}{c}{(n=25)} & \multicolumn{2}{c}{(N=212)} \\
\cmidrule(lr){2-3} \cmidrule(lr){4-5} \cmidrule(lr){6-7} \cmidrule(lr){8-9} \cmidrule(lr){10-11}
Response & n & \% & n & \% & n & \% & n & \% & n & \% \\
\midrule
Yes & 52 & 61.9 & 52 & 85.2 & 37 & 88.1 & 19 & 76.0 & 160 & 75.5 \\
Maybe & 26 & 31.0 & 6 & 9.8 & 2 & 4.8 & 4 & 16.0 & 38 & 17.9 \\
No & 6 & 7.1 & 3 & 4.9 & 3 & 7.1 & 2 & 8.0 & 14 & 6.6 \\
\bottomrule
\end{tabular}
\end{table}

\subsubsection{Willingness to Pay for Improved Services}

A majority of participants (63.3\%) expressed willingness to pay for improved ASR services that better recognize diverse dialects (Table~\ref{tab:q19_pay}). Gulf Coast participants showed the highest willingness (81.5\%), followed by Atlanta (58.8\%), Tucson (56.0\%), and Miami Beach (53.5\%). Among those willing to pay, preferences also varied by site. 

\begin{table}[h]
\centering
\caption{Willingness to Pay for Improved ASR Services by Site (Q19)}
\label{tab:q19_pay}
\begin{tabular}{lcccccccccc}
\toprule
 & \multicolumn{2}{c}{Atlanta} & \multicolumn{2}{c}{Gulf Coast} & \multicolumn{2}{c}{Miami Beach} & \multicolumn{2}{c}{Tucson} & \multicolumn{2}{c}{Total} \\
 & \multicolumn{2}{c}{(n=85)} & \multicolumn{2}{c}{(n=54)} & \multicolumn{2}{c}{(n=43)} & \multicolumn{2}{c}{(n=25)} & \multicolumn{2}{c}{(N=207)} \\
\cmidrule(lr){2-3} \cmidrule(lr){4-5} \cmidrule(lr){6-7} \cmidrule(lr){8-9} \cmidrule(lr){10-11}
Response & n & \% & n & \% & n & \% & n & \% & n & \% \\
\midrule
Prefer 1-time payment & 26 & 30.6 & 10 & 18.5 & 13 & 30.2 & 5 & 20.0 & 54 & 26.1 \\
< \$10/month & 15 & 17.6 & 6 & 11.1 & 5 & 11.6 & 6 & 24.0 & 32 & 15.5 \\
\$10--20/month & 4 & 4.7 & 13 & 24.1 & 4 & 9.3 & 2 & 8.0 & 23 & 11.1 \\
\$20--30/month & 3 & 3.5 & 12 & 22.2 & 1 & 2.3 & 1 & 4.0 & 17 & 8.2 \\
> \$30/month & 2 & 2.4 & 3 & 5.6 & 0 & 0.0 & 0 & 0.0 & 5 & 2.4 \\
Not willing to pay & 35 & 41.2 & 10 & 18.5 & 20 & 46.5 & 11 & 44.0 & 76 & 36.7 \\
\bottomrule
\end{tabular}
\end{table}

\subsection{Qualitative Findings: Lived Experiences Across Four Communities}

Complementing the quantitative results, we present analysis of open-ended narratives which revealed recurring patterns in how users interpret, respond to, and make sense of ASR failures. Rather than simply expressing frustration, participants' accounts exposed the sophisticated strategies they employ to navigate exclusionary systems, the emotional labor required to maintain relationships with failing technologies, and the critical awareness they hold about why these systems fail them. 

\subsubsection{Recognition of Exclusionary Design}
\label{sec:recognition_exclusion}

Across all sites, participants articulated explicit awareness that voice technologies were not designed with their communities in mind. Whereas the causal and technical dimensions of this awareness are examined in \S\ref{sec:critical_awareness}, we focus here on how exclusion is experienced phenomenologically---as affect, identity disruption, and lived certainty derived from repeated failures. A Navajo speaker from Tucson captured this sentiment directly: ``Because Navajo is my first language ... I get this automatic feeling of inferiority if that makes sense. Because ... in my mind I'm thinking this is not something that was made for me automatically. And I'm stepping into a world of what other people are doing and not me because I'm native.'' (Tucson CD24-606). This participant's elaboration reveals how exclusion becomes internalized through accumulated interactions with systems that consistently fail to recognize one's voice. 

A Gulf Coast participant made the systemic nature of this exclusion explicit: ``It's gonna be for everybody and that it needs to be based on everybody's speech and everybody's culture. But I don't take stuff that deep. You know I just look at stuff different, like I know that places weren't created for everyone. Yeah most things aren't inclusive'' (Gulf Coast CD24-546). This statement reveals a paradox: the participant demonstrates sophisticated critical analysis---recognizing that technologies marketed as universal exclude their community---while simultaneously describing this exclusion as unremarkable, something not worth dwelling on because it reflects a broader pattern of non-inclusive design.

An Atlanta participant made the racial dimensions of this exclusion explicit: ``It is very frustrating because for technologies designed to make life better don't work for bi-lingual people of color. Whereas other non-[people of color] around me have no trouble with the technology'' (Atlanta CD24-188). This comparative observation highlights how exclusion becomes visible through contrast with others' seamless experiences.

These statements demonstrate that users do not simply experience ASR failures as random technical glitches. Rather, they understand these failures as evidence of systematic exclusion from the design process, reflecting broader patterns of whose voices, languages, and ways of speaking are centered in technological development.

\subsubsection{Internalized Blame and Emotional Burden}

Despite recognizing systemic exclusion, many participants described internalizing ASR failures as personal inadequacy. This pattern was particularly striking given participants' simultaneous awareness that the problem was not theirs to solve. A Tucson participant articulated this tension: ``It's my problem that this thing's not working. Never. My thought is never like there's something wrong with this you know. Like this needs to be fixed or this is not perfect you know. That thought never entered enters my mind. I'm always like what am I doing wrong. How can I say it to make this thing work you know. And so I think that that's the part that gets most people. Yeah it's kind of like what did I do wrong when it's really not your fault'' (Tucson CD24-606). 

This participant's account reveals the psychological process of misplaced responsibility: despite knowing that the system is flawed, the immediate emotional response is self-blame. The phrase ``what am I doing wrong when it's really not your fault'' captures the gap between the critical understanding and the unavoidable emotional experience. 

Another Tucson participant described feelings of isolation: ``It kinda makes you frustrated actually. ... Like I said, most people don't have problem with it, you know. Like why am I the only one struggling with this'' (Tucson CD24-611). This statement illustrates how ASR failures create the illusion of individual inadequacy in addition to collective exclusion, particularly when failures are not publicly visible or discussed. 

The most succinct expression of this emotional burden came from an Atlanta participant: ``I feel inadequate speaking my own first language'' (Atlanta CD24-198). This statement captures a profound form of linguistic alienation---the internalized sense that one's native language variety, the language of intimate relationships and cultural identity, is somehow deficient because it fails to conform to the narrow linguistic models encoded in voice recognition systems.

\subsubsection{Strategic Accommodation to System Failure}
\label{sec:strategic_accommodation}

Participants across all sites described extensive behavioral modifications required to achieve even basic functionality with voice technologies. These adaptations ranged from phonetic adjustments to code-switching, representing invisible labor that users perform to accommodate systems that do not accommodate them. 

A Gulf Coast participant described the mental and linguistic accommodation required: ``I had to adjust my pronunciation, speaking more slowly and deliberately, almost like I was 'talking to a foreigner.' It worked, but it felt awkward, like I was compromising my natural way of speaking to accommodate the technology'' (Gulf Coast CD24-425). The comparison to ``talking to a foreigner'' reverses the expected relationship---rather than the technology adapting to diverse users, users must treat the technology as a non-native speaker requiring careful and modified input.

Another Gulf Coast participant made the racialized nature of this accommodation explicit: ``Yeah, I put on my white voice when I use it. Like I put on my voice when I talk to Siri, like when I want it to be immediate'' (Gulf Coast CD24-546). This participant's casual reference to deploying a ``white voice'' reveals both the normalization of code-switching as necessary for technological interaction and the explicit awareness that the system recognizes certain racialized varieties of English while failing to recognize others. 

An Atlanta participant expressed the desire for authenticity denied by these constant accommodations: ``Every time I use my natural dialect I have problems. I have learned to adjust my dialect for voice recognition systems. I wish I could just speak normally'' (Atlanta CD24-055). The phrase ``speak normally'' highlights how technological failures redefine speakers' natural, authentic speech as abnormal or problematic. 

A Tucson participant described the cognitive burden of constant self-monitoring: ``I find myself thinking a lot more about how I'm talking than I would in a daily day to day life'' (Tucson CD24-629). This statement reveals how voice technologies introduce extra cognitive load, forcing users to monitor and adjust their speech in real-time rather than focusing on communicative goals.

Participants' accounts suggest that what technical evaluations measure as ``minor performance degradation'' translates, from users' perspectives, into substantial ongoing labor, requiring constant adjustment of linguistic behavior to achieve functionality that other users access seamlessly. 

\subsubsection{Cultural and Linguistic Exclusion}
Participants described specific ways their linguistic and cultural knowledge remains unrecognized by voice technologies, requiring either abandonment of these resources or creation of workarounds. A participant of Guyanese heritage in Miami Beach proposed a solution: "We need a Caribbean Alexa... Because you know in the Caribbean we have our own Creole. And so an Alexa Caribbean is going to know the Caribbean Creole because if you ask Alexa something that is Caribbean based she doesn't have that culture nor ethnicity that she can relate to" (Miami Beach CD24-569). This participant's proposal anthropomorphizes the problem effectively—the system lacks not just linguistic data but cultural knowledge and context necessary for genuine comprehension.

A Miami Beach participant described how accent becomes marked as non-normative: ``When I call someone I have to use non-Miami accent, normal people talk'' (Miami Beach CD24-600). The equation of ``non-Miami accent'' with ``normal people talk'' reveals how ASR systems encode particular varieties as standard while rendering others abnormal, creating a hierarchy of acceptable speech. 

Another Gulf Coast participant identified the technical gap: ``Siri doesn't take into account different dialects (ex. AAVE, Ebonics, French accents, slang)'' (Gulf Coast CD24-503), naming specific varieties and features excluded from the system's model of language. 

A Gulf Coast participant described the specific linguistic features that remain unrecognized: ``Voice to text doesn't recognize or know what to do with common slang terms like 'chile.' I have had issues where voice to text doesn't understand when I am trying to play a song that has a title that is slang. If I do not speak super proper Siri doesn't understand me. Most of my names with black cultural roots aren't understood when I ask Siri to call them so Siri will call the wrong person'' (Gulf Coast CD24-515). This account catalogs the cumulative exclusions---slang terms, song titles reflecting cultural tastes, personal names---that together render a person's linguistic and social world invisible to the system.

\subsubsection{Critical Awareness of Systemic Failures}
\label{sec:critical_awareness}

Beyond the experiential recognition of exclusion described in \S\ref{sec:recognition_exclusion}, many participants moved from describing how exclusion feels to diagnosing why it occurs, locating problems in inadequate training data, narrow linguistic models, or design processes that exclude diverse communities. A Gulf Coast participant analyzed the technical limitation: ``I will realize that it is not intelligent enough, and it needs to constantly improve and update the language habits and database'' (Gulf Coast CD24-495). This participant identifies insufficient training data as the source of failures rather than accepting user blame. 

Some participants contextualized ASR failures within broader technological exclusion. The Gulf Coast participant who noted ``It's gonna be for everybody but it's not based on everybody'' (Gulf Coast CD24-546, also quoted in \S\ref{sec:recognition_exclusion}) moved beyond describing how exclusion feels to articulating a structural critique: that claimed universality masks particular, exclusionary design choices.

This critical awareness did not necessarily translate into expectations that systems would not change. As one participant noted: ``I know that it will get better'' (Gulf Coast CD24-537), expressing faith in future improvement that allows acceptance of present failures. Another stated: ``it's easy to feel frustrated but sometimes I just have to tell myself to, like you know, relax, you know, it's learning'' (Miami Beach CD24-558), framing the persistent failures as temporary growing pains rather than unsolvable problems. 

\section{Discussion}

\subsection{The Expectations-Experience Gap}

Over 80\% of participants expected ASR systems to understand their speech at least 75\% of the time, yet more than half reported that technologies fail to consider their cultural backgrounds. This gap between expectations and experience has significant implications. Marketing narratives that promise ``natural'' voice interaction may create expectations that systems will accommodate linguistic diversity. When systems fail, users who have internalized these promises may attribute failures to their own speech rather than to system limitations. The participant who described always thinking ``what am I doing wrong'' despite knowing ``it's really not your fault'' (Tucson CD24-606) captures this dynamic precisely. 

This expectations gap also explains the substantial rates of technology abandonment we observed. When systems consistently fail to meet reasonable expectations, users face a choice between continued accommodation labor and discontinuation. That 80.0\% of Tucson participants and 55.8\% of Miami Beach participants reported stopping use of ASR technologies suggests that for many users, the costs of accommodation eventually outweigh the benefits of use. 

\subsection{The Paradox of Critical Awareness and Internalized Blame}

One of the most significant findings is the coexistence of sophisticated critical awareness with internalized self-blame. Participants articulated clear understanding that systems were ``not made for'' them, that training data excludes their communities, and that their failures reflect design choices rather than personal deficiencies. Yet this critical awareness did not prevent emotional responses of inadequacy, frustration directed inward, and self-monitoring of speech. 

This paradox extends theoretical work on intersectional invisibility \cite{bhattacharyya_you_2023} into human-technology interaction. Bhattacharyya and Berdahl found that ambiguous invisibility experiences (erasure, whitening) elicit shame and self-blame even when structural causes are recognized. Our findings suggest similar dynamics operate with ASR: repeated system failures create cumulative experiences of technological invisibility that generate internalized responses despite intellectual understanding of systemic causes. The participant who described feeling ``inadequate speaking my own first language'' (Atlanta CD24-198) exemplifies how technological exclusion can create profound linguistic alienation.

\subsection{Invisible Labor and Unequal Burden}

The coping strategies participants reported---repetition (82.8\%), slowing speech (71.3\%), speaking louder (72.9\%), enunciating more (66.7\%)---represent substantial invisible labor. This labor is invisible in two senses: it goes unrecognized in technical evaluations that measure only accuracy, and it remains unseen by users for whom systems ``just work.'' 

The participant who described deploying a ``white voice'' when using Siri (Gulf Coast CD24-546) makes explicit what remains implicit in aggregate statistics: from participants' perspectives, accommodation to ASR systems requires performing linguistic identities that align with dominant speech norms while suppressing natural speech patterns. This linguistic accommodation---the invisible labor defined in \S\ref{sec:tech_labor}---represents a form of everyday discrimination in which access to technology is contingent on abandoning authentic self-expression \cite{lippi-green_english_2012}. 

Notably, this burden falls unequally even among marginalized communities. Tucson participants (primarily Native American speakers) reported the highest rates of challenges across all categories and the highest rates of technology abandonment (80.0\%), suggesting that speakers of Indigenous languages and varieties who rarely feature in accessibility and fairness scholarship face particularly severe exclusion from voice technologies. 

\subsection{Implications for Participatory System Development}

The finding that 75.5\% of participants expressed willingness to contribute voice samples despite experiencing significant harm raises important questions for participatory approaches to system development. An Atlanta participant who described feeling ``embarrassed, sad, and defeated'' by ASR failures---and who reported that the technology ``makes me feel like I don't speak well or struggle with basic English''---nevertheless expressed willingness to contribute voice samples, adding: ``if y'all reach out later, I would be open to doing it'' (Atlanta CD24-163). Yet this same participant insisted ``I feel I should be compensated,'' rejecting the premise that improving a system that has caused harm should come at further personal cost. This response encapsulates a tension that ran through our data: willingness to contribute coexists not with na\"ive optimism but with clear-eyed recognition that such contributions constitute labor deserving fair exchange.

Participants' motivations for this willingness appeared to vary. Some expressed altruistic investment in future users: the same Atlanta participant stated ``I just want the device to work for people like me'' (CD24-163), locating motivation in collective benefit rather than personal gain. Others framed continued engagement as pragmatic patience---``I know that it will get better'' (Gulf Coast CD24-537)---or active emotional management: ``it's easy to feel frustrated but sometimes I just have to tell myself to, like you know, relax, you know, it's learning'' (Miami Beach CD24-558). This last statement is particularly revealing: the participant performs emotional labor (in Hochschild's \cite{hochschild_managed_2012} sense) to sustain engagement with a system that repeatedly fails them, reframing persistent exclusion as a temporary developmental stage.

This willingness to contribute represents an additional layer of invisible labor layered atop the accommodation burden documented in \S\ref{sec:strategic_accommodation}. Communities already bearing involuntary costs---code-switching, repetition, self-monitoring---are asked to take on voluntary labor to fix the systems imposing those costs. That 75.5\% expressed willingness while a majority (63.3\%) were also willing to pay for improved services underscores the depth of investment these communities maintain in voice technologies despite their exclusionary design. These findings support calls for design justice frameworks \cite{costanza-chock_design_2020} that center affected communities not merely as data sources but as partners with authority over development processes. The willingness our participants expressed should not be treated as a free resource to be extracted; it should be met with equitable compensation, shared decision-making, and accountability for the harms that generated the need for improvement in the first place. 

\subsection{Cross-site Comparison}
The communities sampled in this study (Atlanta, Tucson, the Gulf Coast, and Miami Beach) and the ethnicities targeted within these (African American, Native American, Mississippi French/French Creole and Hispanic) were selected for their distinct histories, relationships with other communities in the region, and socio-cultural experiences.  We expected that cross-community differences would emerge, and this proved to be true.  The African American and Native American communities showed greatest awareness of the burden to code-switch--both for themselves and their communities. Speakers in these two groups also tended to display the highest levels of individual meta-linguistic awareness (e.g., explaining how they made stylistic adjustments away from their home dialect). Ongoing research in our laboratory further explores the possibility of an association between the burden to code-switch and experiences of invisibility. On the other hand, the Miami Beach speakers tended to comment on the experiences of peers rather than make comments that placed them within their broader community. 

As was discussed above, all three of these groups discussed discontinuing the use of tech that did not satisfy their expectations. The group of which we had uncertain predictions was the Gulf Coast group, which is the least-studied of the speech communities included in this research. We were interested to find that these participants expressed the most self-blame. Although recognizing the failure of technology to work consistently for them, they were the group most likely to display acceptance and continued use.

\section{Conclusions}

This study demonstrates that understanding ASR bias requires moving beyond accuracy metrics to examine lived experiences of technological exclusion. Our findings reveal that critical awareness of systemic bias coexists with feelings of invisibility and internalized blame: participants who clearly articulated that systems were not made for them nevertheless described feeling inadequate speaking their native languages. This paradox, knowing the system is at fault while still blaming oneself, suggests that intellectual understanding provides incomplete protection against the psychological toll of repeated technological rejection.

The accommodation strategies participants reported represent substantial invisible labor that current fairness metrics fail to capture. Code-switching, hyper-articulation, repetition, and constant self-monitoring constitute an unequal tax on users whose speech falls outside narrow training distributions. Access to voice technology is often contingent on suppressing authentic linguistic identity. This burden falls most heavily on already marginalized communities. 

Our study contributes to growing recognition that technical fairness metrics alone cannot capture the full dimensions of algorithmic harm. Voice technologies create worlds that include some speakers while excluding others, and the invisible costs of exclusion extend far beyond what error rates can measure. Yet our findings also reveal grounds for optimism: despite experiencing significant harm, participants overwhelmingly expressed willingness to contribute to model improvement and pay for improved services. This finding suggests that at least some in affected communities remain invested in shaping voice technologies rather than abandoning them entirely. Realizing this potential requires moving beyond extraction-based approaches that treat marginalized speakers as data sources toward participatory frameworks that recognize community expertise, ensure equitable compensation, and grant meaningful authority over how systems are developed and deployed. The path toward inclusive voice technologies runs through the communities currently excluded from them. 

%%
%% The acknowledgments section is defined using the "acks" environment
%% (and NOT an unnumbered section). This ensures the proper
%% identification of the section in the article metadata, and the
%% consistent spelling of the heading.
\begin{acks}
Thanks are first due to the people who were willing to share their time and experiences with us. We would like to thank the members of the Seattle Black Community Advisory Board (B-CAB), and the students of the University of Washington Human-Centered Design and Engineering working group led by Jay L. Cunningham for feedback on survey design and for community partnership. Finally, we thank the SocioLinx team at the University of Washington Department of Linguistics for their work in the community, in the lab, and in months of vigorous conversation: Didar Akar, Devyn Brandt, Chase Fossen, Ty Gill-Saucier, Florian Humbert, Didem Ikizoğlu, Dayton Kelly, Mwinkuma Kuunapor, Paige Morris, and Amina Venton.
\end{acks}

%%
%% The next two lines define the bibliography style to be used, and
%% the bibliography file.
\bibliographystyle{ACM-Reference-Format}
\bibliography{sample-base}

%%
%% If your work has an appendix, this is the place to put it.

\appendix

\section{Questionnaire Design}

\subsection{Complete Questionnaire}

The following presents the complete questionnaire administered to participants. Question types are indicated as follows: [O] = Open-ended response; [MC] = Multiple choice; [MA] = Multiple selection (select all that apply); [Likert] = 5-point Likert scale. The Atlanta site used an earlier version of the survey with some different response options, as noted below.

\noindent\textbf{Demographics and Speech Background}

\textbf{Q1} [O]: How would you describe the way that you speak?

\textbf{Q2} [O]: Would you be willing to share how you self-identify in terms of ethnicity or cultural background? If so, what term(s) do you use?

\textbf{Q3} [MC]: Please indicate your age range: \textit{Under 30 / 30--50 / Over 50}

\textbf{Q4} [O]: Do you have any speech challenges (impairments or disabilities) e.g. stuttering, stammering, slurring, etc. that might affect your ability to use voice recognition technologies effectively? Please describe.

\noindent\textbf{Voice Technology Use}

\textbf{Q5} [MA]: You are now going to see some different types of Small and Large Vocabulary technologies. Please select the ones that you use or have used. \textit{Options: Stand-alone Smart Speakers (e.g., Amazon Echo, Google Home, Siri HomePod); Virtual Assistants on Mobile Phones, Tablets or Smart TVs; Voice Commands in Smart Home Appliances; Voice Commands in Car Infotainment Systems; Voice-controlled Gaming Devices; Voice Biometrics for Authentication; Voice-enabled Wearable Devices}

\textbf{Q6a} [O]: If you selected any of the Small or Large Vocabulary uses above: What are some tasks you use this tech for?

\textbf{Q6b} [O]: If there are any other tasks you use your Small or Large Vocabulary devices for, please specify.

\textbf{Q7} [MA]: You are now going to see some different types of Conversational technologies. Please select the ones that you use or have used. \textit{Options: Video conferencing apps with auto-generated transcription (e.g., captions and transcripts on Zoom, Microsoft Teams, etc.); Voice-to-text/dictation features; Automated phone systems (IVR)}

\textbf{Q8a} [O]: If you selected any of the Conversational uses above: What are some tasks you use your tech for?

\textbf{Q8b} [O]: If there are any other tasks you use your Conversational devices for, please specify.

\textbf{Q9} [O]: If there are any other kinds of voice recognition technologies you have used, please specify.

\textbf{Q10} [O]: Are there any devices from the list we just talked about that you used, but stopped using? Can you tell us about what led you to stopping?

\noindent\textbf{Experiences with Voice Recognition}

\textbf{Q11} [O]: Can you tell us about your most memorable experience with voice recognition technology? What was the technology and how did you use it? Did you accomplish what you wanted through using it?

\textbf{Q12a--f} [Likert]: How often do you use voice recognition technology for the following purposes? \textit{Scale: 1 = Very frequently to 5 = Never.} (a) To be productive and manage tasks; (b) Using voice commands as shortcuts for accessing apps on my device; (c) To keep up with information, news, weather; (d) To play around and discover new capabilities of voice technology; (e) To support me in accessibility (captioning speech); (f) Other (please specify).

\noindent\textbf{Dialect and Accent Challenges}

\textbf{Q13} [O]: Can you share a time where you had a challenge with your way of speaking (dialect, accent or language) being understood by voice recognition systems?

\textbf{Q14} [MA]: Which, if any of the following reasons describe the challenge(s) that you identified above, or other challenges that you have experienced? Select all that apply.

\textit{Options for the Gulf Coast, Miami Beach, and Tucson:} Voice technologies don't seem to take into consideration my cultural/ethnic/racial background; Voice assistants take what I say out of context and either don't work or do something I didn't want; It takes too much physical effort for me to speak clear enough to be understood; I feel self-conscious about how I speak when talking to voice technologies

\textit{Additional options for Atlanta only:} Voice technologies force me to change the way I talk to be understood; Voice technologies force me to repeat my speech over and over to be understood; When voice assistants/smart devices don't recognize my speech, I still have to manually complete a task; Voice technologies don't work for my accent/dialect most or all of the time; Voice technologies don't seem useful for my day-to-day lifestyle

\noindent[Skip logic prompts for narrative elicitation]: \textit{You indicated that voice technologies don't consider your background. Would you like to share a time this happened? / You indicated that you feel self-conscious using voice tech. Would you like to share a time you felt this way? / You indicated that sometimes you get unexpected results using voice tech. Would you like to share a time this happened?}

\textbf{Q14b} [O]: If there are any other challenges or issues you have experienced, please specify.

\noindent\textbf{Coping Strategies}

\textbf{Q15a} [MA]: If you have experienced challenges in using voice recognition, are there things you do to fix or deal with the challenge? Select all that apply.

\textit{Options for Gulf Coast, Miami Beach, and Tucson:} Repeat what I said; Change the words or phrases I use; Slow down my speech; Enunciate more; Speak louder; Imitate the voice assistant; Get another person to control the device for me; Manually complete the task instead; I just stop using it

\textit{Options for Atlanta:} Repeat what I said; Change the way that I speak/code-switch; Imitate the voice assistant; Get another person to control the device for me; I just stop using it

\textbf{Q15b} [O]: If you said above that there are other ways you fix or deal with challenges in using voice recognition, please specify.

\noindent\textbf{Emotional Responses}

\textbf{Q16} [O]: How does it make you feel when your voice assistant doesn't understand your speech/makes frequent mistakes?

\noindent\textbf{Expectations and Willingness}

\textbf{Q17} [MC]: How often do you expect voice recognition technologies to be able to understand your speech dialect or accent correctly? \textit{Options: Always (100\% every time I use my device); Often (75\% --- most times I use my device); Sometimes (50\% --- every now and again); Rarely (25\% occasionally); Never (0--20\%)}

\textbf{Q18a} [MC]: Would you be willing to provide voice samples or feedback to help improve voice recognition technologies' accuracy for different dialects and accents? \textit{Options: Yes, I would be interested in contributing; Maybe, I would need more information; No, I would not be interested in contributing}

\textbf{Q19} [MC]: Finally, would you be willing to pay for a product/service that provides improved voice recognition for diverse dialects and helps in undoing bias? If yes, which describes how much you'd be willing to pay? \textit{Options: Not willing to pay---I feel I should be compensated; Prefer 1-time payment; < \$10/month; \$10--20/month; \$20--30/month; More than \$30/month}

\noindent\textbf{Voice Sample Collection (Optional)}

\textbf{Q20}: Would you like to contribute your voice to our database? [If yes, participants were directed to a voice recording interface.]

\textbf{Q21} [O]: Please share any thoughts or comments on participating in this exercise. What questions might you have for our team regarding the study?

\section*{Generative AI Usage Statement}

ChatGPT 5.2 was used to assist with grammar editing and table formatting. The authors retain full responsibility for the originality, accuracy, and integrity of the manuscript.

\end{document}